\documentclass[11pt]{article} 
\usepackage{rldmsubmit,palatino}
\usepackage{graphicx}

\renewcommand\footnotemark{}

\title{An Analysis of Measure-Valued Derivatives for Policy Gradients}

\author{
Jo\~{a}o Carvalho
\thanks{
An extended version of this work can be found at \href{https://arxiv.org/pdf/2107.09359.pdf}{https://arxiv.org/pdf/2107.09359.pdf}.
}
\\
Department of Computer Science\\
Technische Universit\"{a}t Darmstadt, Germany\\
\texttt{joao@robot-learning.de} \\
\And
Jan Peters
\\
Department of Computer Science\\
Technische Universit\"{a}t Darmstadt, Germany\\
\texttt{jan.peters@tu-darmstadt.de} \\
}

\usepackage{array}  
\usepackage{graphicx}  
\usepackage{amsmath, bm}  
\usepackage{amsthm}   
\usepackage{amsfonts} 
\usepackage{amssymb} 
\usepackage{mathtools}
\usepackage{glossaries}

\usepackage[ruled,vlined,linesnumbered]{algorithm2e}

\usepackage{subfig}
\usepackage[export]{adjustbox}

\usepackage[inline, shortlabels]{enumitem}

\usepackage{mathtools}

\usepackage{wrapfig}

\usepackage{hyperref}

\usepackage{dblfloatfix}

\renewcommand{\vec}[1]{\bm{#1}}
\newcommand{\mat}[1]{\mathbf{#1}}

\newcommand{\policy}{\pi}

\newcommand{\statespace}{\mathcal{S}}
\newcommand{\actionspace}{\mathcal{A}}

\newcommand{\state}{\vec{s}}
\newcommand{\nextstate}{\vec{s}'}
\newcommand{\action}{\vec{a}}
\newcommand{\omegavec}{\vec{\omega}}
\newcommand{\thetavec}{\vec{\theta}}
\newcommand{\phivec}{\vec{\phi}}

\newcommand{\xvec}{\vec{x}}
\newcommand{\epsvec}{\vec{\varepsilon}}

\newcommand{\dpartialomegak}{\partial/\partial \omega_k}

\newcommand*\de{\mathop{}\!\mathrm{d}}

\newcommand{\wrt}{w.r.t.}

\DeclarePairedDelimiterX{\infdivx}[2]{(}{)}{%
	#1\;\delimsize\|\;#2%
}

\newcommand{\bigO}[1]{\mathcal{O}\!\left(#1\right)}

\newcommand{\R}{\mathbb{R}}
\newcommand{\E}[2]{\mathbb{E}_{#1}\left[#2\right]}

\newcommand{\V}[2]{\mathbb{V}_{#1}\left[#2\right]}
\newcommand{\cov}[2]{\mathrm{Cov}_{#1}\left[#2\right]}

\newcommand{\transpose}{\intercal}

\newcommand{\covariance}{\vec{\Sigma}}

\newcommand{\Gaussian}[1]{\mathcal{N}\left(#1\right)}

\newcommand{\Ptrans}{\mathcal{P}}

\newcommand{\policyparams}{\thetavec}

\newcommand{\gradpolicyparams}{\nabla_{\thetavec}}
\newcommand{\policyparametrized}{\policy_{\policyparams}}
\newcommand{\distributionparams}{\omegavec}
\newcommand{\distributionparamk}{\omega_k}
\newcommand{\graddistributionparams}{\nabla_{\omegavec}}
\newcommand{\graddistributionparamk}{\nabla_{\omega_k}}

\glsdisablehyper
\newacronym{ml}{ML}{Machine Learning}
\newacronym{rl}{RL}{Reinforcement Learning}
\newacronym{mdp}{MDP}{Markov Decision Process}

\newacronym{reptrick}{Rep-trick}{Reparametrization trick}
\newacronym{mvd}{MVD}{Measure-Valued Derivative}
\newacronym{sf}{SF}{Score-function}

\newacronym{vae}{VAE}{Variational Autoencoder}
\newacronym{vi}{VI}{Variational Inference}
\newacronym{elbo}{ELBO}{Evidence Lower Bound}

\newacronym{mc}{MC}{Monte Carlo}

\newacronym{fd}{FD}{Finite-difference}

\newacronym{lqr}{LQR}{Linear-Quadratic Regulator}

\newacronym{pgpe}{PGPE}{Policy Gradients with Parameter based Exploration}
\newacronym{nes}{NES}{Natural Evolutionary Strategies}
\newacronym{rwr}{RWR}{Reward Weighted Regression}
\newacronym{reps}{REPS}{Episodic Relative Entropy Policy Search}
\newacronym{nac}{NAC}{Natural Actor Critic}
\newacronym{power}{PoWER}{Policy Learning by Weighting Exploration with the Returns}
\newacronym{expected-sarsa}{Expected SARSA}{Expected SARSA}

\newacronym{trpo}{TRPO}{Trust Region Policy Optimization }
\newacronym{ppo}{PPO}{Proximal Policy Optimization}
\newacronym{ddpg}{DDPG}{Deep Deterministic Policy Gradient}
\newacronym{td3}{TD3}{Twin Delayed DDPG}
\newacronym{sac}{SAC}{Soft Actor-Critic}
\newacronym{gae}{GAE}{Generalized Advantage Estimation}

\newacronym{gpomdp}{}{GPOMPDP}
\newacronym{reinforce}{}{REINFORCE}

\newacronym{emvd}{E-MVD}{Episodic-MVD}
\newacronym{sacmvd}{SAC-MVD}{Soft Actor-Critic with MVD}
\newacronym{treemvdpg}{Tree-MVD}{Tree-MVD-Policy Gradient}

\newacronym{sacextrasamples}{SAC-extra-samples}{}

\newacronym{sacsf}{SAC-SF}{}
\newacronym{sacsfextrasamples}{SAC-SF-extra-samples}{}

\newacronym{spsa}{SPSA}{Simultaneous Perturbation Stochastic Approximation }

\begin{document}

\maketitle

\begin{abstract}
Reinforcement learning methods for robotics are increasingly successful due to the constant development of better policy gradient techniques.
A \textit{precise} (low variance) and \textit{accurate} (low bias) gradient estimator is crucial to face increasingly complex tasks.
Traditional policy gradient algorithms use the likelihood-ratio trick, which is known to produce unbiased but high variance estimates.
More modern approaches exploit the reparametrization trick, which gives lower variance gradient estimates but requires differentiable value function approximators.
In this work, we study  a different type of stochastic gradient estimator - the Measure-Valued Derivative. This estimator is unbiased, has low variance, and can be used with differentiable and non-differentiable function approximators.
We empirically evaluate this estimator in the actor-critic policy gradient setting and show that it can reach comparable performance with methods based on the likelihood-ratio or reparametrization tricks, both in low and high-dimensional action spaces.
With this work, we want to show that the Measure-Valued Derivative estimator can be a useful alternative to other policy gradient estimators.
\end{abstract}



\startmain 

\section{Introduction}

Complex robotics tasks, such as locomotion and manipulation, can be formulated as \gls{rl} problems in continuous state and action spaces~\cite{Sutton2018IntroRL}.
In these settings, the \gls{rl} objective is commonly an expectation dependent on a parameterized policy, whose parameters are optimized via gradient ascent with the policy gradient theorem~\cite{sutton1999PG}.
As this gradient cannot always be obtained in closed-form, there are three main approaches to obtain an unbiased estimate: the likelihood-ratio, also called \gls{sf}; the Pathwise Derivative, commonly known as the \gls{reptrick}; and the \gls{mvd}~\cite{mohamed2019monte}.
The \gls{sf} has been use extensively in policy gradients, but is known to produce high-variance estimates.
REINFORCE is an example of a \gls{sf} based method~\cite{williamsReinforce1992}.
Generally, the \gls{reptrick}~\cite{kingma2014autoencoding} is a low variance gradient estimator, but it requires the optimized function to be differentiable, which excludes non-differentiable approximators, such as regression trees
. 
\gls{sac}~\cite{Haarnoja2018SAC} is an example of an algorithm that uses this estimator.
\gls{mvd} is a third method to compute unbiased gradient estimates, 
which is not commonly used in the \gls{ml} community.
It generally has low variance and unlike the \gls{sf} avoids computing an extra baseline for variance reduction.
Few works have studied its application to \gls{rl}~\cite{bhatt2019pgweak,krishnamurthy2011realtime}. 
In this work, we analyze the properties of \glspl{mvd} in actor-critic policy gradient algorithms in the \gls{lqr} problem, for which we know the true policy gradient, and with an off-policy algorithm for complex environments with high-dimensional
spaces.
The results suggest that this estimator can be an alternative to compute policy gradients in continuous action spaces.

\section{Background}
\label{sec:background}

\subsection{Reinforcement Learning and Policy Gradients}
\label{section:reinforcementlearning}
Let a \gls{mdp} be defined as a tuple $\mathcal{M} = (\statespace, \actionspace, \mathcal{R}, \mathcal{P}, \gamma, \mu_0)$, where $\statespace$ is a continuous state space $\state \in \statespace$, $\actionspace$ is a continuous action space $\action \in \actionspace$, $\mathcal{P}: \statespace \times \actionspace \times \statespace \to \R$ is a transition probability function, with $\mathcal{P}(\nextstate | \state, \action)$ the density of transitioning to state $\nextstate$ when taking action $\action$ in state $\state$, $\mathcal{R}: \statespace \times \actionspace \to \R$ is a reward function, $\gamma \in [0, 1) $ is a discount factor, and $\mu_0:\statespace \to \R$ the initial state distribution.
A policy $\policy$ defines a (stochastic) mapping from states to actions. 
The discounted state distribution under a policy $\policy$ is given by $d^\policy_\gamma(\state) = (1-\gamma) \sum_{t=0}^{\infty} \gamma^t P(\state_t = \state | \state_0, \policy)$, where $\state_0$ is the initial state, and $P: \statespace \to \R$ the probability of being in state $\state$ at time step $t$.
The state-action value function is the discounted sum of rewards collected from a state-action pair following the policy $\policy$, $Q^{\policy}(\state, \action) = \E{\policy,\Ptrans }{\sum_{t=0}^\infty \gamma^t r(\state_t, \action_t) | \state_0 = \state, \action_0 = \action}$, and the state value function is its expectation \wrt~the policy $V^{\policy}(\state) = \E{\action \sim \policy}{Q^{\policy}(\state, \action)}$. 
The advantage function is the difference between the two $A^{\policy}(\state, \action) = Q^{\pi}(\state, \action) - V^{\policy}(\state)$.
The goal of a \gls{rl} agent is to maximize the expected sum of discounted rewards from any initial state
$J(\policy) = \E{\state_0 \sim \mu_0}{V^{\policy}(\state_0)}$.
In continuous action spaces, it is common to optimize $J$ using gradient ascent with the policy gradient theorem~\cite{sutton1999PG}
$\gradpolicyparams J(\policyparametrized) = \E{\state \sim \mu_{\gamma}^{\policy}}{\int \gradpolicyparams \policy(\action | \state; \policyparams) Q^\policy(\state, \action) \de \action}$.
Note that the integral over the action space is not an expectation. 


\subsection{Monte Carlo Gradient Estimators}
\label{section:mc-gradient-estimation}
The objective function of several problems in \gls{ml}, e.g. \gls{vi}, is often posed as an expectation of a function $f$ \wrt~a distribution $p$ parameterized by $\distributionparams$,
$J(\distributionparams) = \E{p(\xvec; \distributionparams)}{f(\xvec)} = \int p(\xvec; \distributionparams) f(\xvec) \de \xvec$,
where $f: \R^n \to \R$ is an arbitrary function of $\xvec\in\mathbb{R}^n$, $p:\R^n \times \R^m \to \R$ is the distribution of $\xvec$, and $\distributionparams \in \R^m$ are the parameters encoding the distribution - also known as distributional parameters.
For instance, in amortized \gls{vi}
~\cite{kingma2014autoencoding}
, $f$ is the \gls{elbo}, and if $p$ is a multivariate Gaussian that approximates the posterior distribution, $\distributionparams$ aggregates the mean and covariance.
A gradient ascent algorithm is a common choice to find the parameters that maximize $J(\distributionparams)$, for which we need to compute 
$\graddistributionparams J(\distributionparams) = \int \graddistributionparams p(\xvec; \distributionparams) f(\xvec) \de \xvec$.
Since the derivative of a distribution is in general not a distribution itself, the integral is not an expectation and thus not solvable directly by \gls{mc} sampling.
If $f$ also depends on $\distributionparams$, the product rule is applied.
Let a \gls{mc} estimation of the gradient be obtained as
$\widehat{\graddistributionparams J(\distributionparams)} \approx  1/M \sum_{i=1}^{M} \hat{\vec{g}}_i$, with $\hat{\vec{g}}_i \in \R^m$ an unbiased gradient estimate.
There are three known ways to build an unbiased estimator of $\graddistributionparams J(\distributionparams)$ - \gls{reptrick}, \gls{sf} and \gls{mvd}.
Due to space constraints, we refer the reader to~\cite{mohamed2019monte} for more details on the \gls{reptrick} and \gls{sf} estimators.

\textbf{\glsfirst{mvd}}~\cite{Pflug89}
Even though in general the derivative of a distribution \wrt~its parameters is not a distribution, it is a difference between two distributions up to a normalizing constant~\cite{Pflug89}.
The main idea behind \gls{mvd} is to write the derivative \wrt~a single distributional parameter $\distributionparamk \in \R$ as a difference between two distributions, i.e. $\graddistributionparamk p(\xvec; \distributionparams) = c_{\distributionparamk} \left( p^+_{\distributionparamk}(\xvec; \distributionparams) - p^-_{\distributionparamk}(\xvec; \distributionparams) \right)$, where $c_{\distributionparamk}$ is a normalizing constant that can depend on $\distributionparamk$, and $p^+_{\distributionparamk}$ and $p^-_{\distributionparamk}$ are two distributions referred as the positive and negative components, respectively. 
The \gls{mvd} is the triplet $ \left(c_{\distributionparamk}, p^+_{\distributionparamk}(\xvec; \distributionparams), p^-_{\distributionparamk}(\xvec; \distributionparams) \right)$. 
For common distributions, such as the Gaussian, Poisson, or Gamma, decompositions have been analytically derived~\cite{mohamed2019monte}.
The \gls{mvd} decomposition gives $\graddistributionparamk J(\distributionparams) = \int c_{\distributionparamk} \left(p^+_{\distributionparamk}(\xvec; \distributionparams) - p^-_{\distributionparamk}(\xvec; \distributionparams) \right) f(\xvec) \de \xvec$.
I.e., the derivative is the difference of two expectations scaled with the normalization constant.
For example, to compute the derivative of a one-dimensional Gaussian $\mathcal{N}(x; \mu, \sigma)$ \wrt~$\sigma$, we can sample the positive part from a Double-sided Maxwell and the negative one from a Gaussian.
\begin{wrapfigure}{h}{0.22\textwidth}
    \vspace{-0.3cm}
	\centering
	\subfloat[Quadratic\label{fig:grad-test-functions-quadratic}]{%
		\includegraphics[width=0.15\textwidth,valign=t]{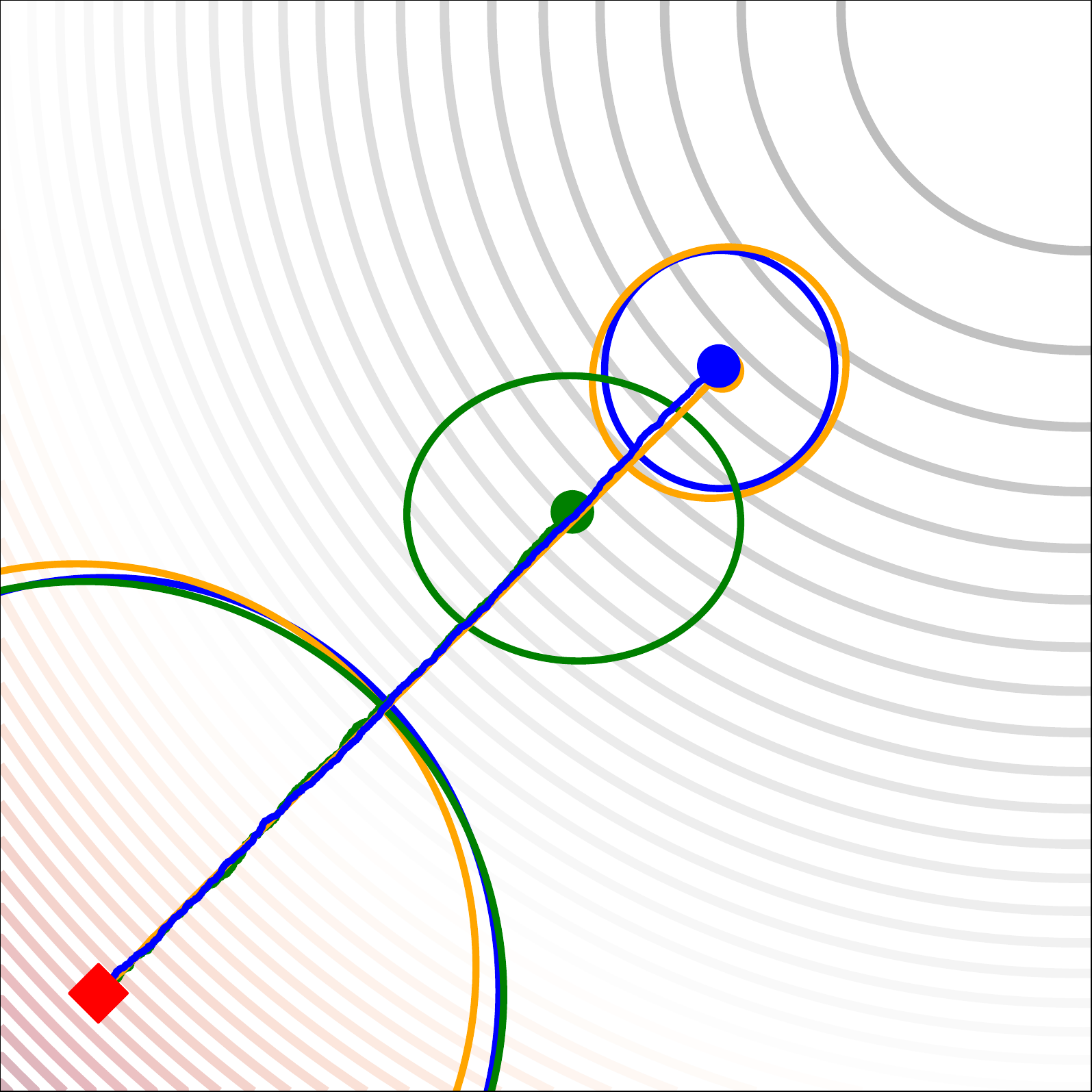}}
    \\
	\subfloat[Styblinski]{%
		\includegraphics[width=0.15\textwidth,valign=t]{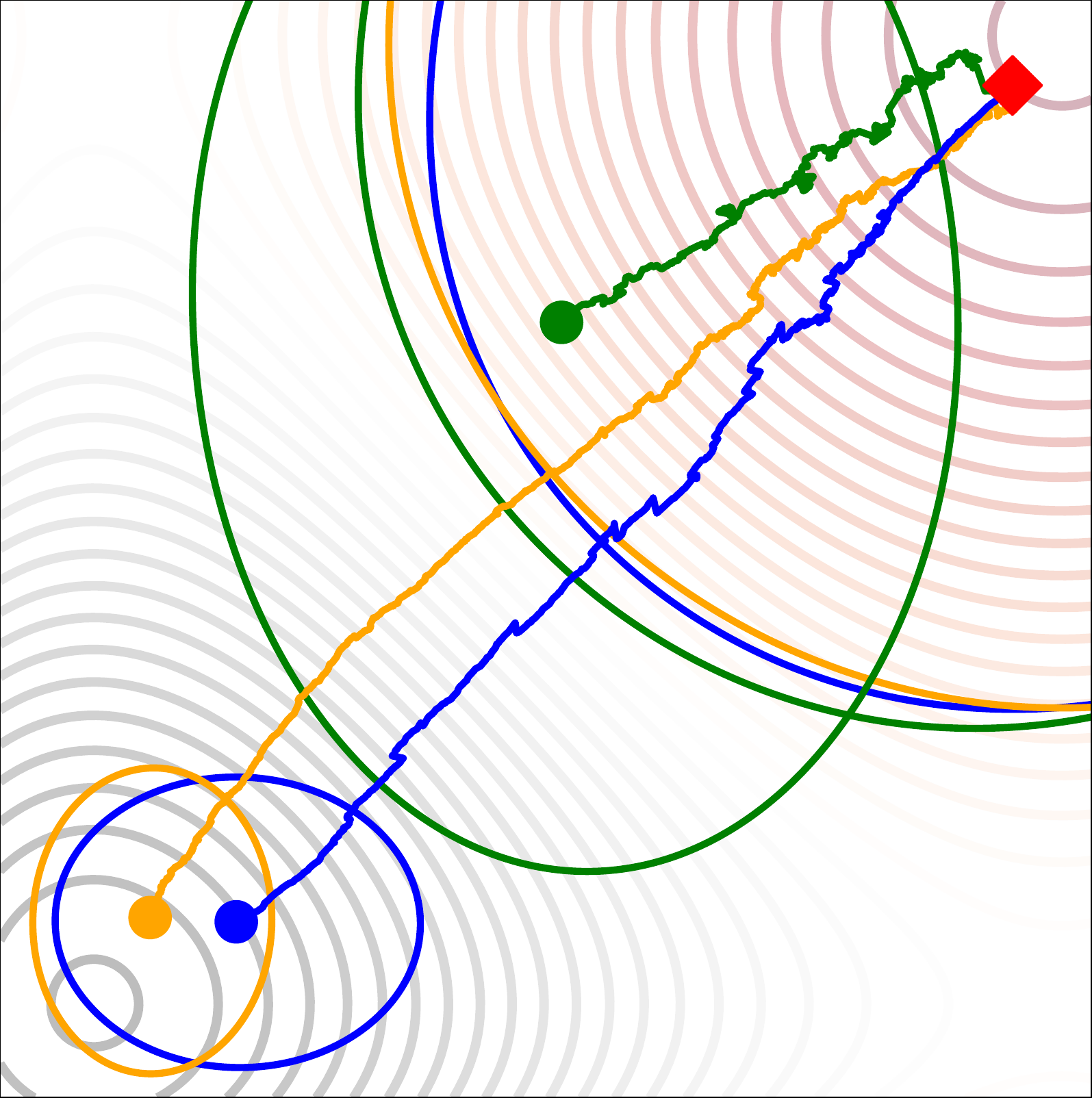}}
	\\
	\includegraphics[width=0.22\textwidth]{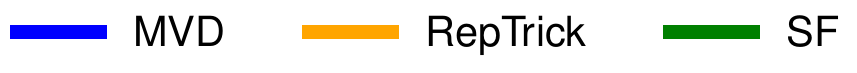}
	\caption{Sample run of the unbiased gradient estimators in test functions. Lines show the means of the distributions, and ellipses one standard deviation at the beginning (diamond) and end (circle).}
	\label{fig:grad-test-functions} 
	\vspace{-1.8cm}
\end{wrapfigure}
The gradient \wrt~all parameters is the concatenation of the derivatives \wrt~all individual parameters $\distributionparamk$, which results in $\bigO{2|\distributionparams|}$ queries to $f$ to compute the full gradient.
In contrast, the \gls{sf} and \gls{reptrick} have $\bigO{1}$ complexity.
The variance of this estimator is
$\V{p(\vec{x};\distributionparams)}{\hat{g}^{\textrm{MVD}}_k}  = \V{p^+_{\distributionparamk}(\vec{x};\distributionparams)}{f(\vec{x})} + \V{p^-_{\distributionparamk}(\vec{x};\distributionparams)}{f(\vec{x})} - 2 \cov{p^+_{\distributionparamk}(\vec{x};\distributionparams)p^-_{\distributionparamk}(\vec{x}';\distributionparams)} {f(\vec{x}), f(\vec{x}')}$
~\cite{mohamed2019monte},
which depends on the chosen decomposition and how correlated are the function evaluations at the positive and negative samples.
If $p$ is a multivariate distribution with independent dimensions it factorizes as $p(\xvec; \distributionparams) = \prod_{i=1}^{n} p(x_i; \bm{\xi}_i)$, with $x_i \in \R$ and $\bm{\xi}_i$ a subset of $\distributionparams$
(if $p(\xvec; \distributionparams)$ is a Gaussian with diagonal covariance, then ${\bm{\xi}_i = \{\mu_i, \sigma_i\}}$).
The derivative \wrt~one parameter $\distributionparamk \in \bm{\xi}_k$ is given by
$\dpartialomegak p(\xvec; \distributionparams) = c_{\distributionparamk} \left( p^+_{\distributionparamk}(x_k; \bm{\xi}_k) - p^-_{\distributionparamk}(x_k; \bm{\xi}_k) \right) \prod_{i=1, i \neq k}^{n} p(x_i; \bm{\xi}_i) =  c_{\omega_k} \left( p^+_{\distributionparamk}(\xvec; \distributionparams) - p^-_{\distributionparamk}(\xvec; \distributionparams) \right)$
, where $p^+_{\distributionparamk}(\xvec; \distributionparams)$ is the original multivariate distribution with the $k$-th component replaced by the positive part of the decomposition of the univariate marginal $p(x_k; \bm{\xi}_k)$ \wrt~$\omega_k$ (the negative component is analogous).

\vspace{-0.3cm}
\paragraph{Illustrative Example}
Fig.~\ref{fig:grad-test-functions} shows the optimization of the expectation of common test functions \wrt~a 2-dimensional Gaussian distribution with diagonal covariance, using gradient ascent.
All estimators use $8$ \gls{mc} samples per gradient update.
The \gls{sf} uses an optimal baseline for variance reduction.
\gls{reptrick} and \gls{mvd} consistently move towards the global maximum, while the \gls{sf} shows unstable behaviors. 

\section{Actor-Critic Policy Gradients with Measure-Valued Derivatives}
\label{sec:stepbased_mvd}

Given the properties of \glspl{mvd}, especially its low variance, we propose to analyze them in actor-critic policy gradients, which can be seen as computing an unbiased stochastic gradient estimate of the expected return.
Let $\policy(\action | \state; \distributionparams = g(\state; \policyparams))$ be a stochastic policy, where $\distributionparams$ are distributional parameters resulting from applying $g$ with parameters $\policyparams$ to the state $\state$, e.g. neural networks that output the mean and covariance of a Gaussian distribution.
Since the gradient of $g$ \wrt~$\policyparams$ can be easily computed if $g$ is a continuous deterministic function, we only consider the gradient \wrt~$\distributionparams$.
In Table \ref{tab:policygradient} we write the policy gradient theorem for a single parameter $\distributionparamk$ with the three different estimators.
The \gls{mvd} formulation of the policy gradient shows that to compute the gradient of one distributional parameter we can sample from the discounted state distribution by interacting with the environment, and then evaluate the $Q$-function at actions sampled from the positive and negative components of the policy decomposition conditioned on the sampled states, $\policy_{\distributionparamk}^+(\cdot | \state; \distributionparams)$ and $\policy_{\distributionparamk}^-(\cdot | \state; \distributionparams)$, respectively.
Importantly, the $Q$-function estimate is the one from $\policy$ and not from $\policy^+$ or $\policy^-$.
The function approximator for $Q$ can be a differentiable one, such as a neural network, or a non-differentiable one, such as a regression tree
.
For the \gls{sf} estimator, the $Q$-function is replaced by the advantage function
, which keeps the estimator unbiased but has lower 
variance.

The \gls{mvd} formulation assumes we can query the $Q$-function for the same state with multiple actions.
While previous work~\cite{bhatt2019pgweak} assumed access to a simulator that can estimate the return for different actions starting from the same state, e.g. with \gls{mc} rollouts, this scenario is not generally applicable to \gls{rl}, where the agent cannot easily reset to an arbitrary state, especially in real-world environments.
Hence, we assume a critic approximator is available, e.g. fitted from samples.

\begin{table*}[hb]
	\centering
	\begin{tabular}{ l l } 
		\textbf{Score-Function} & $\graddistributionparamk J(\distributionparams) = \E{\state \sim \mu^{\pi}_{\gamma}, \; \action \sim \policy(\cdot | \state; \distributionparams)}{\graddistributionparamk \log \policy(\action | \state; \distributionparams) Q^\policy(\state, \action)}$ \\
		\textbf{Reparametrization Trick} & $\graddistributionparamk J(\distributionparams) = \mathbb{E}_{\state \sim \mu^{\pi}_{\gamma}, \; \epsvec \sim p_{\epsvec}}\left[ \nabla_{\action} Q^\policy(\state, \action = h(\state, \epsvec; \distributionparams)) \graddistributionparamk h(\state, \epsvec; \distributionparams) \right] $ \\
		\textbf{Measure-Valued Derivative} &
		$\graddistributionparamk J(\distributionparams) = \mathbb{E}_{\state \sim \mu^{\pi}_{\gamma}} \left[ c_{\distributionparamk} \left( \E{\action \sim \policy_{\distributionparamk}^+(\cdot | \state; \distributionparams)}{Q^\policy(\state, \action)} -  \E{\action \sim \policy_{\distributionparamk}^-(\cdot | \state; \distributionparams)}{Q^\policy(\state, \action)} \right) \right]$
	\end{tabular}
	\caption{Policy gradient theorem with the different unbiased stochastic gradient estimators.}
	\label{tab:policygradient}
	\vspace{-0.3cm}
\end{table*}

\subsection{Gradient Analysis in the Linear Quadratic Regulator}
\label{sec:lqr_analysis}

\begin{figure*}[t]
	\noindent
	\centering
	\hspace{-0.5cm}
	\begin{minipage}[b]{0.333\textwidth}
		\subfloat[Gradient errors with true $Q$]{
			\includegraphics[scale=0.58]{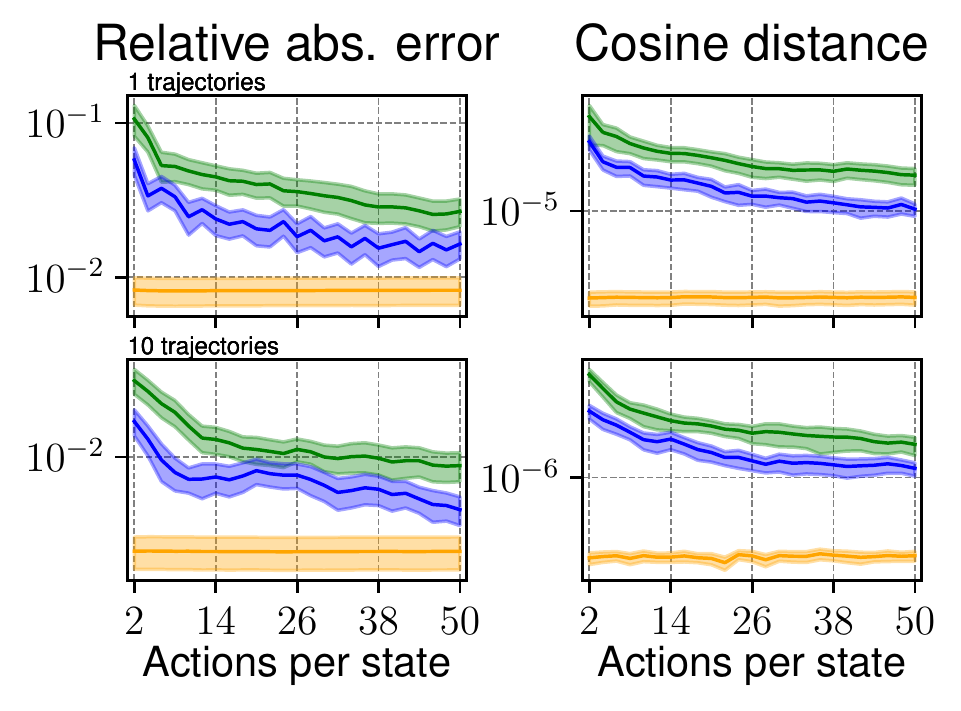} 
			\label{fig:lqr-results-graderror-trueq} 
		}
	\end{minipage}
	\hspace{-0.87cm}
	\begin{minipage}[b]{0.333\textwidth}
		\subfloat[Gradient errors with approx. $Q$]{
			\includegraphics[scale=0.7]{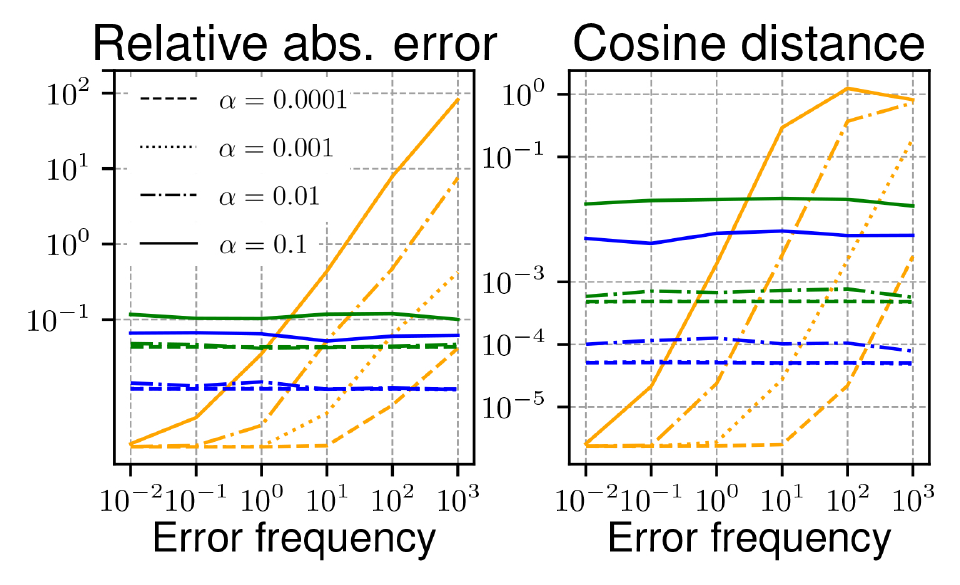}
			\label{fig:lqr-results-graderror-approxq} 
		}
	\end{minipage}
	\hspace{0.3cm}
	\begin{minipage}[b]{0.333\textwidth}
		\subfloat[Learning curves with approx. $Q$]{
			\includegraphics[scale=0.57]{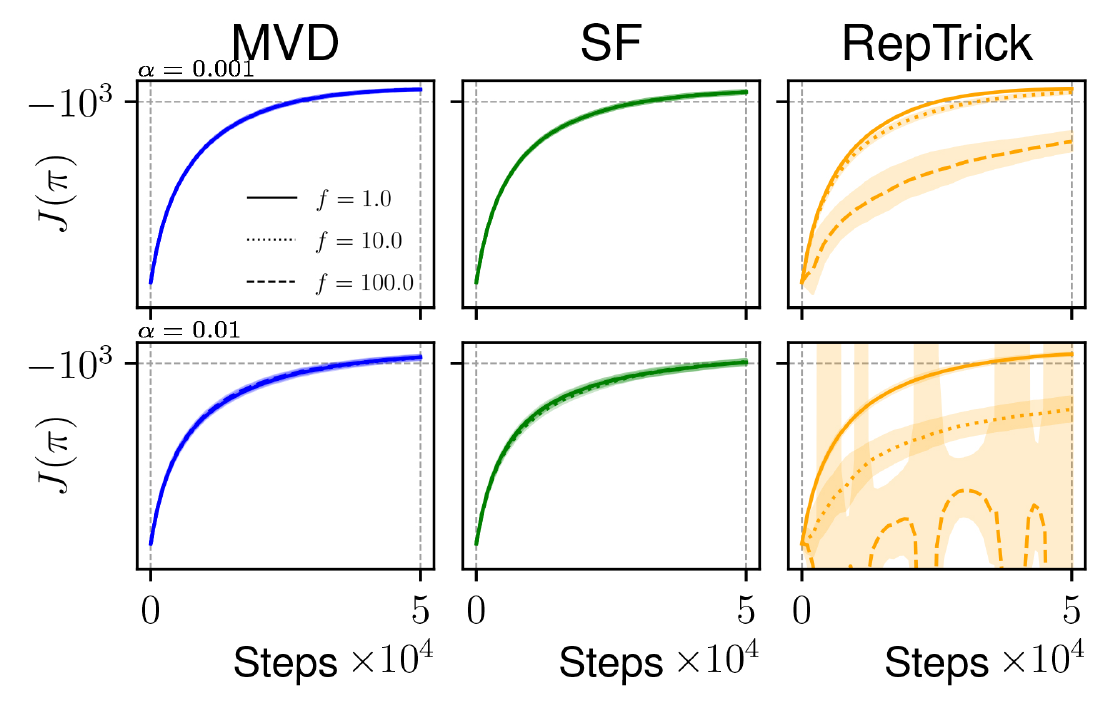}
			\label{fig:lqr-results-learning-curve} 			
		}
	\end{minipage}
	\\
	\includegraphics[width=0.29\linewidth,valign=t]{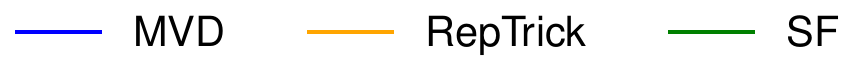}
	\caption{
	Gradient analysis for a LQR with $2$ states and $1$ action.
	a) Gradient errors in magnitude and direction for $1$ and $10$ trajectories (rows) and sampled actions.
	b) Gradient errors with increasing error noise and frequency. 
	The different linestyles correspond to different error amplitudes $\alpha$. 
	The results are estimated using $10$ trajectories and $20 |\actionspace|$ actions per state.
	c) Learning curves for different error amplitudes (rows) and frequencies. The gradients are estimated using $1$ trajectory and $2|\mathcal{\actionspace}|$ actions per state. 
	Some configurations lead to indistinguishable results, and thus plots appear superposed.
	The solid lines depict the mean and the shaded area the $95\%$ confidence interval of 25 random seeds.
	}
	\label{fig:lqr-results} 
	\vspace{-0.5cm}
\end{figure*}

We consider a discounted infinite-horizon discrete-time \gls{lqr} with a linear Gaussian stochastic policy $\policy(\action|\state) = \Gaussian{\action | -\mat{K}\state, \covariance} $, where $\mat{K} \in \R^{|\actionspace|\times|\statespace|}$ is a learnable feedback gain matrix, and $\covariance \in \R^{|\actionspace|\times|\actionspace|}$ a fixed covariance. 
Since the value function is known and can be computed by numerically solving the Algebraic Riccati Equation, as well as the gradient \wrt~$\mat{K}$, the \gls{lqr} is a good baseline for policy gradient algorithms.
We construct an \gls{lqr} with $|\statespace|=2$ states and $|\actionspace|=1$ action, with dynamics such that the uncontrolled system is unstable, and select randomly a suboptimal initial gain $\mat{K}_{\mathrm{init}}$ such that the closed-loop is stable.
The initial state is uniformly drawn and fixed for all rollouts.
We compare the policy gradient of the expected return \wrt~$\mat{K}_{\mathrm{init}}$ with the estimators from Table~\ref{tab:policygradient}.
For the \gls{sf}, we replace the $Q$-function with the advantage function for variance reduction.
The expectation of the discounted on-policy state distribution is sampled by interaction with the environment, but the expectation over actions uses the LQR's true critics $Q$ and $V$, to represent an idealized scenario.

We study two sources of errors -  
the \textit{relative absolute error} $ \left(| \|\hat{\vec{g}} \| - \| \vec{g} \| | \right) / \| \vec{g} \| $ that relates the magnitude of the estimated $\hat{\vec{g}}$ and true gradient $\vec{g}$,
and the \textit{cosine distance} $1 - \hat{\vec{g}}^\transpose\vec{g}/(\|\hat{\vec{g}} \| \| \vec{g} \|) $ that is the direction error.
An ideal estimator has both errors close to zero.
The errors are analyzed along two dimensions -- the number of trajectories and the number of actions sampled to solve the action expectations.
The results from Fig.~\ref{fig:lqr-results-graderror-trueq} show that, as expected, with the number of trajectories fixed, increasing the number of sampled actions decreases the estimators' errors.
The \gls{reptrick} achieves the best results in both magnitude and direction in all environments, and the \gls{mvd} is slightly better than the \gls{sf}.
In our experiments we observed this holds for higher dimensions as well.
This reveals that even though the \gls{sf} complexity is $\bigO{1}$, in practice we need roughly the same number of samples as \gls{mvd} to obtain a low error gradient estimate.
Knowing that the $Q$-function is quadratic in the action space, the results are in line with the example from Fig.~\ref{fig:grad-test-functions-quadratic} where a quadratic function was optimized and all estimators performed equally well.
Next, we consider a scenario where the true $Q$-function has to be estimated.
For that we model the approximator with a local approximation error on top of the true value as $\hat{Q}(\state,\action) = Q(\state,\action) + \alpha Q(\state,\action)\cos(2\pi f \vec{p}^\transpose \vec{a} + \phi)$,
i.e. we add noise proportional to the true estimate, where $\alpha$ represents the fraction of the true $Q$ amplitude, $f$ is the error frequency, $\vec{p}$ is a random vector sampled from a symmetric Dirichlet distribution, and $\phi\sim\mathcal{U}[0, 2\pi]$ a phase shift.
$\vec{p}$ and $\phi$ introduce randomness to remove correlation between action dimensions.
Fig.~\ref{fig:lqr-results-graderror-approxq} shows the gradient errors in magnitude and direction as a function of the error frequency and amplitude.
The error frequency does not affect the gradient estimation of \gls{sf} or \gls{mvd},
as can be seen by the horizontal blue and green lines. 
On the contrary the \gls{reptrick} is heavily affected by both.
This result is in line with the theory, since under a Gaussian distribution the variance of the \gls{reptrick} is upper-bounded by the Lipschitz constant of the derivative of $Q$~\cite{mohamed2019monte}. 
Fig.~\ref{fig:lqr-results-learning-curve} shows the expected discounted return per steps taken in the environment with different amplitudes and frequencies of errors. 
The \gls{sf} and \gls{mvd} do not suffer from errors in the estimation and converge towards the optimal policy even in the presence of high-frequency error terms.
The \gls{reptrick} on the other hand either shows slower convergence or fails to converge due to the poor gradient estimates.
From these experiments we observe that even though the \gls{reptrick} provides the most precise and accurate gradient under a true value function, this does not always hold for approximated functions, especially when there is an action correlated error.


\subsection{MVDs in Off-Policy Policy Gradient for Deep Reinforcement Learning}
\label{sec:mvd_deeprl}

\begin{figure*}[t] 
	\centering
	\includegraphics[width=0.86\textwidth,valign=t]{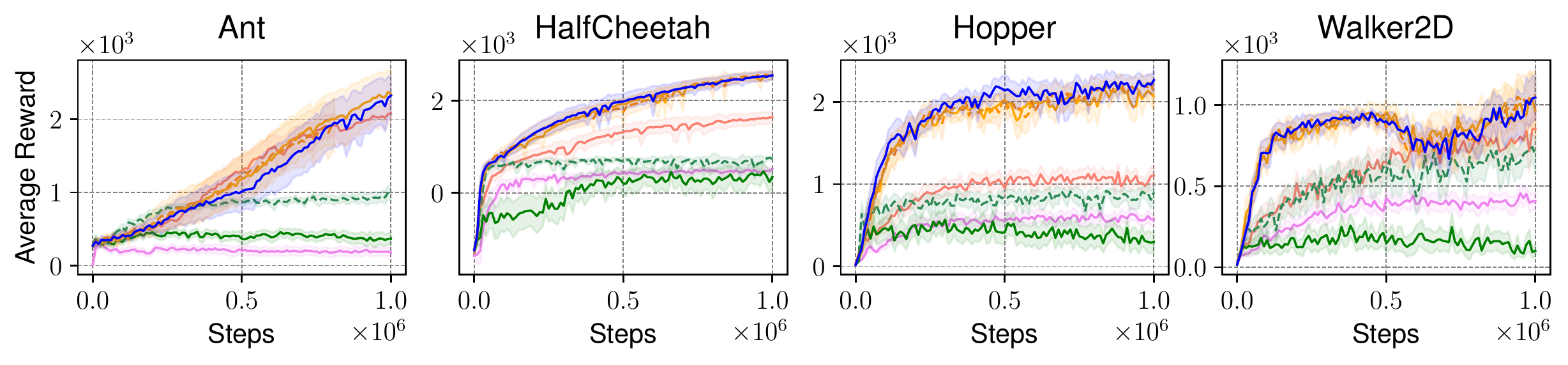}
	\\
	\includegraphics[width=1.0\textwidth]{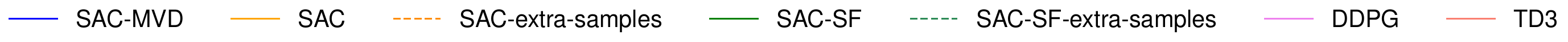}
	\caption{Policy evaluation results during training on different tasks in deep \gls{rl}. The lines depict the mean of the average reward per samples collected and the shaded area the $95\%$ confidence interval of 25 random seeds.}
	\label{fig:step-based-experiments}
	\vspace{-0.4cm}
\end{figure*}

Next we illustrate how \glspl{mvd} can be used in a deep \gls{rl} algorithm such as \gls{sac}, whose
surrogate objective is
${J_{\policy}(\distributionparams) = \E{\state \sim d^{\beta}, \action \sim \policy(\cdot | \state; \distributionparams)}{Q^{\policy}(\state, \action; \phivec) -\alpha \log \policy (\action | \state ; \distributionparams)  } = \E{\state \sim d^{\beta}, \action \sim \policy(\cdot | \state; \distributionparams)}{f(\state, \action; \phivec, \distributionparams)}}$,
where $d^{\beta}$ is an off-policy state distribution, $Q$ is a neural network parameterized by $\phivec$, $\alpha$ weighs the entropy regularization term, and $\distributionparams$ are the distributional parameters of $\policy$.
\gls{sac} estimates the gradient \wrt~$\distributionparams$ using the \gls{reptrick}.
Instead we propose to compute it using \gls{mvd} and name this modification as \acrshort{sacmvd},
whose gradient \wrt~a single parameter $\distributionparamk$ becomes
${\graddistributionparamk J_{\policy}(\distributionparams) = \mathbb{E}_{\state \sim d^{\beta}} \left[ c_{\distributionparamk} \left( \mathbb{E}_{\action \sim \policy_{\distributionparamk}^+(\cdot | \state; \distributionparams)} \left[ f(\state, \action; \phivec, \distributionparams) \right] -  \mathbb{E}_{\action \sim \policy_{\distributionparamk}^-(\cdot | \state; \distributionparams)} \left[f(\state, \action; \phivec, \distributionparams) \right] \right) - \mathbb{E}_{\action \sim \policy(\cdot | \state; \distributionparams)} \left[ \graddistributionparams \alpha \log \policy (\action | \state ; \distributionparams) \right] \right]}$,
where $\policy_{\distributionparamk}^+$ and $\policy_{\distributionparamk}^-$ are the positive and negative components of the \gls{mvd} decomposition. 

We benchmark \acrshort{sac} with the different gradient estimators in high-dimensional continuous control tasks from the PyBullet simulator
with: \acrshort{sacmvd} with one \gls{mc} sample; \acrshort{sacsf} - a version of \acrshort{sac} using the \acrshort{sf}; \acrshort{sacsfextrasamples} - same as \acrshort{sacsf} but with the same number of queries as \acrshort{sacmvd} per gradient estimate; \acrshort{sacextrasamples} - same as \acrshort{sac} but with the same number of gradient estimates as \acrshort{sacsfextrasamples};
\acrshort{ddpg}~\cite{Lillicrap2016DDPG} and \acrshort{td3}~\cite{fujimoto2018td3} - two off-policy algorithms.
The average reward curves obtained during training are shown in Fig.~\ref{fig:step-based-experiments}.
\gls{sac} with \gls{reptrick} and \acrshort{sacmvd} show similar performance, and increasing the number of \gls{mc} samples does not improve the overall results (\acrshort{sacextrasamples} performs equally well).
Even though \acrshort{sacmvd} needs more forward passes of the $Q$-function, \gls{sac} needs to backpropagate through it, and so we found the computation time to be similar in our implementation.
The results of our experiments reveal that the \gls{reptrick} is not fundamental for the performance of \gls{sac}, and suggest that the superior performances of this algorithm depend on other aspects, such as the entropy regularization, the state-dependent covariance and the squashed Gaussian policy.
Additionally, it is worth noticing that \glspl{mvd} are applicable to function classes where the \gls{reptrick} is not, which allows to explore other approximator classes while using the benefits of \gls{sac}.


\section{Conclusion}
\label{sec:conclusion}

We presented \glspl{mvd} as an alternative to the \gls{sf} and \gls{reptrick} estimators for actor-critic policy gradient algorithms.
The empirical results showed that methods based on the \gls{mvd} are a viable alternative to the other two, and differently from~\cite{bhatt2019pgweak}, we avoided resetting the environment to a specific state, showing how \glspl{mvd} are applicable to the general \gls{rl} framework.
In the simple \gls{lqr} environment and with an oracle critic, the \gls{mvd} performs better than the \gls{sf} and worse than the \gls{reptrick}.
However, with an action dependent error, the \gls{mvd} and \gls{sf} estimates are not affected by the error frequency, while the \gls{reptrick} is sensitive to it.
In tasks with high-dimensional action spaces we obtain comparable results with the \gls{reptrick} using only one gradient estimate, which shows that it is not crucial for the \gls{sac} algorithm. 
Furthermore, unlike in \gls{sac}, \glspl{mvd} do not require a differentiable $Q$-function, allowing the use of other types of function approximators.
In future work we will investigate how to reduce the computational complexity of \glspl{mvd} by computing derivatives along important dimensions and using a convex combination the estimators, which still remains unbiased~\cite{mohamed2019monte}.



\bibliography{references.bib}
\bibliographystyle{IEEEtran}

\end{document}